\documentclass[runningheads]{llncs}
\usepackage{amsmath}
\usepackage{cleveref}
\usepackage[T1]{fontenc}
\usepackage{cite}
\usepackage{graphicx}

\begin{document}
\sloppy
\title{Probing Length Generalization in Mamba via Image Reconstruction}
\author{Jan Rathjens\inst{1}\orcidID{0000-0002-0490-538X} \and
Robin Schiewer\inst{2}\orcidID{0000-0002-4021-2476} \and
Laurenz Wiskott\inst{1}\orcidID{0000-0001-6237-740X} \and
Anand Subramoney\inst{3}\orcidID{0000-0002-7333-9860}}
%
% \author{Anonymous submission}
\authorrunning{J. Rathjens et al.}
% \institute{Anonymous submission}

\institute{
Institute for Neural Computation, Faculty of Computer Science, \\Ruhr University Bochum, Bochum, Germany\\
\email{jan.rathjens@rub.de, laurenz.wiskott@rub.de}
\and
CITEC, Bielefeld University, Bielefeld, Germany\\
\email{robin.schiewer@uni-bielefeld.de}
\and
Department of Computer Science, \\
Royal Holloway, University of London, Egham, UK\\
\email{anand.subramoney@rhul.ac.uk}
}

\maketitle              % typeset the header of the contribution
\begin{abstract}
Mamba has attracted widespread interest as a general-purpose sequence model due to its low computational complexity and competitive performance relative to transformers. However, its performance can degrade when inference sequence lengths exceed those seen during training. We study this phenomenon using a controlled vision task in which Mamba reconstructs images from sequences of image patches. By analyzing reconstructions at different stages of sequence processing, we reveal that Mamba qualitatively adapts its behavior to the distribution of sequence lengths encountered during training, resulting in strategies that fail to generalize beyond this range. To support our analysis, we introduce a length-adaptive variant of Mamba that improves performance across training sequence lengths. Our results provide an intuitive perspective on length generalization in Mamba and suggest directions for improving the architecture.

\end{abstract}

\section{Introduction}
Mamba~\cite{gu_mamba_2023} is a general-purpose sequence model within the class of state space models (SSMs). Compared to the ubiquitous sequence model transformer~\cite{vaswani_attention_2017}, Mamba offers reduced computational complexity while maintaining competitive performance across a broad range of benchmarks. The model has therefore seen widespread interest and adoption, particularly in natural language processing~(NLP)~\cite{gu_mamba_2023,dao_transformers_2024} but also in other domains such as vision~\cite{zhu_vision_2024,liu_vmamba_2024}. 

A central property of sequence models in general is length generalization, i.e., the ability of a model trained on sequences up to a certain length to maintain performance when evaluated on longer sequences. While the original work on Mamba reported promising length generalization capabilities~\cite{gu_mamba_2023}, subsequent studies~\cite{ben-kish_decimamba_2024, ye_longmamba_2025,azizi_mambaextend_2025,lu_mamba_2025,ruiz_understanding_2025,chen_stuffed_2024} have shown that Mamba often fails to generalize to longer sequences across a wide range of NLP tasks.

In this work, we investigate Mamba's length generalization capabilities in a controlled vision-domain setting. Specifically, we train Mamba models to reconstruct Omniglot~\cite{lake_human-level_2015} characters from sequences of image patches of varying lengths. By generating reconstructions at different stages of sequence processing, we obtain indirect but interpretable evidence about the architecture's processing mechanisms. Our setup enables targeted probing experiments to analyze these mechanisms systematically. We further compare Mamba's performance with that of an analogous transformer baseline.

% Our results indicate that Mamba qualitatively adapts its computational strategy to the distribution of sequence lengths encountered during training, and the resulting strategy does not generalize reliably to substantially shorter or longer sequences. Rather than learning length-agnostic dynamics, Mamba learns a compromise across the range of lengths observed during training, effectively trading off performance across encountered sequence lengths. To probe this mechanism, we introduce a sequence-adaptive variant of Mamba, which yield improved performance across the spectrum of training sequence lengths. Additionally, we find that length generalization is task-dependent and that transformers are substantially more robust when exposed to longer sequences.

% Taken together, our work provides a more intuitive and mechanistic understanding of length generalization in Mamba and offers guidance for the design of future variants aimed at improving robustness across varying sequence lengths.

\section{Related Work}
\label{sec:related_work}
Related work on length generalization in Mamba typically follows a common pattern: Studies report substantial limitations on an NLP task, identify architectural properties that may restrict length generalization, and propose targeted modifications. These modifications are then shown empirically to improve length generalization.

For instance, DeciMamba~\cite{ben-kish_decimamba_2024} and LongMamba~\cite{ye_longmamba_2025} argue that Mamba learns an effective receptive field implicitly bounded by training sequence lengths. To mitigate this limitation, DeciMamba discards tokens deemed uninformative based on the magnitude of $\mathrm{\Delta}_t$, Mamba’s input-dependent discretization parameter that modulates hidden-state evolution based on the values of sequence tokens. LongMamba similarly utilizes $\mathrm{\Delta}_t$ but discards tokens only for channels that are believed to encode global sequence information.

Attributing length generalization failures to out-of-distribution (OOD) accumulation of $\mathrm{\Delta}_t$ for long sequences, MambaExtend~\cite{azizi_mambaextend_2025} downscales this value at each time step. Similarly, MambaModulation~\cite{lu_mamba_2025} scales the state transition matrices based on the hypothesis that OOD growth or decay of state norms causes degradation when evaluating on longer sequences.

Taking a broader perspective, Ruiz and Gu~\cite{ruiz_understanding_2025} argue that long sequences induce OOD hidden-state distributions because training explores only a limited subset of attainable states. To mitigate this mismatch, they initialize the hidden state during training with the final state of a preceding sequence, thereby exposing the model to a broader range of state configurations.

Lastly, focusing on Mamba-2~\cite{dao_transformers_2024}, Chen et al.~\cite{chen_stuffed_2024} attribute failures in length generalization to an overparameterized hidden state that hinders learning an effective forgetting mechanism for earlier tokens. They further argue that training sequence length should be balanced with hidden-state size.

While prior work offers valuable analyses of Mamba’s length generalization failures, the diversity of proposed explanations suggests that the underlying causes remain unclear. To complement these efforts, we investigate length generalization in a controlled vision setting and probe Mamba’s internal dynamics through interpretable visualizations.

\section{Method}
\label{sec:method}
To generate interpretable visualizations during sequence processing, we consider an image reconstruction/completion task on $128\times128$ grayscale images of Omniglot characters~\cite{lake_human-level_2015}. Their structured, sparse strokes are easier to interpret than those of natural images.

\begin{figure}[h] 
    \centering 
    \includegraphics[width=1.0\linewidth]{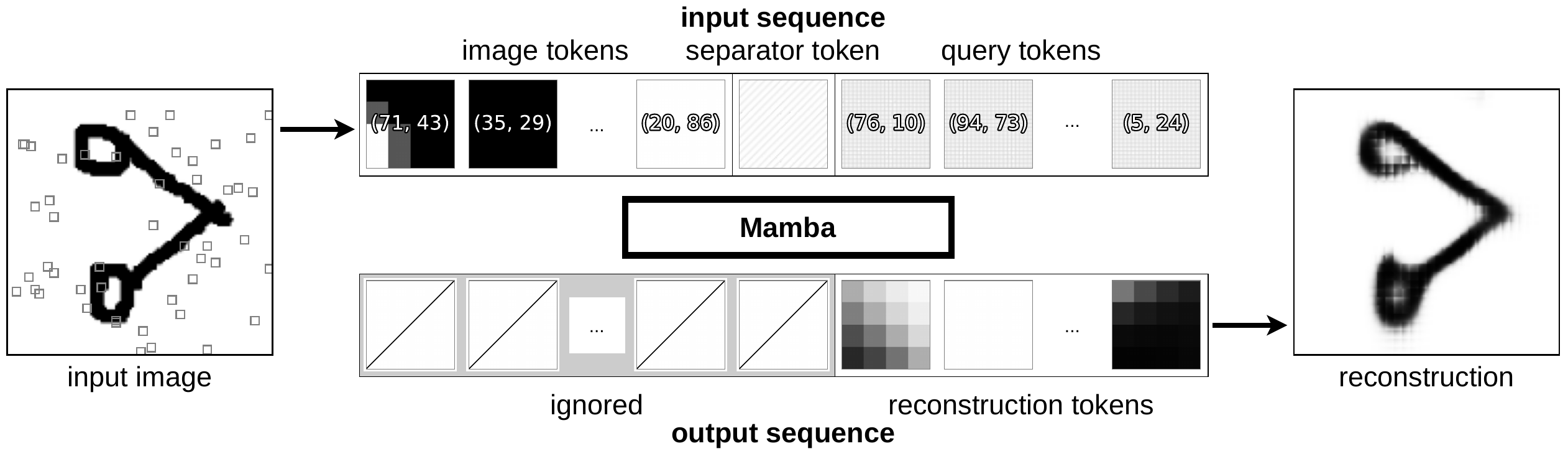}
    \caption{Model overview. An input sequence of image patches augmented with spatial information, a separator token, and query tokens specifying spatial locations is processed by a Mamba model. The model produces reconstruction tokens, each predicting an image patch for the queried location. When query tokens densely cover the image, the reconstruction tokens can be assembled into a visualization of the input image.} 
    \label{fig:method} 
\end{figure}

An overview of the processing pipeline is shown in \Cref{fig:method}. From each image, we sample an arbitrary number of potentially overlapping $4\times4$ patches at random spatial locations. Each patch is flattened and linearly projected to a 14-dimensional embedding vector.

To encode spatial information, we concatenate normalized $(x,y)$ coordinates of the patch center to its embedding, yielding a 16-dimensional representation. We treat these representations as a sequence of tokens and refer to them as image tokens. We also evaluated an alternative positional encoding in which patches were embedded into 16-dimensional tokens and fixed sinusoidal encodings were added following standard transformer practice~\cite{vaswani_attention_2017}. However, this configuration yielded slightly inferior performance compared to coordinate concatenation.

To the sequence of image tokens, we append a learnable 16-dimensional separator token that indicates the end of the image context. We then append a set of query tokens, each encoding a spatial location at which image information is requested. Specifically, a query token is obtained by applying a linear projection to the normalized $(x,y)$ coordinates of the target location, resulting in a 16-dimensional representation. Target locations may coincide with sampled patches but may also lie at arbitrary positions within the image, in which case the model must interpolate the underlying image content.

We refer to the concatenation of image tokens, the separator token, and query tokens as the input sequence. This sequence is processed by a Mamba model, which produces an output token for each input token. The output tokens are passed through a lightweight linear prediction head that preserves dimensionality. We discard outputs corresponding to image and separator tokens and retain only those corresponding to query tokens. These tokens represent predicted image patches at the queried locations and are referred to as reconstruction tokens. When query tokens densely cover the spatial extent of the image, the reconstruction tokens can be optionally assembled into a full image reconstruction for visual inspection. For the $128\times128$ images used in our experiments, 1024 query tokens suffice to cover the entire image. Beyond enabling visualization during processing, our setup also enables analyzing two distinct sequence-processing tasks: incorporating information from image tokens and responding to query tokens.

We train a standard Mamba architecture. Unless stated otherwise, our default configuration consists of eight Mamba blocks with model dimension 16 and state size 8, providing a favorable trade-off between reconstruction performance and computational cost.
Our implementation builds on the Hugging Face Transformers library~\cite{wolf_huggingfaces_2019}. The only architectural modification is the removal of the depthwise convolution over the four most recent tokens at the beginning of each Mamba block. This operation is unnecessary in our setting because random patch sampling removes meaningful local ordering, increases computational overhead, and did not measurably affect performance.

Training is performed on the Omniglot training dataset with random rotation, translation, and scaling augmentations applied on the fly. For each batch, we sample a random number of image patches between $1$ and a parameterized maximum $T_I$ (default $T_I=1024$). In addition, we use a fixed number of query tokens $T_Q$ (default $T_Q=1024$) to produce the output. Unlike image tokens, the number of query tokens can remain fixed because a reconstruction loss can be computed iteratively for each query token, which is not the case for image tokens.

The training objective minimizes the mean squared error (MSE) between reconstruction tokens and the ground-truth image patches at their query locations. Models are trained using standard deep learning techniques on NVIDIA A100 and H100 GPUs for approximately 16,000 epochs. Detailed hyperparameters and implementation specifics are provided in the accompanying code repository\footnote{\url{https://github.com/wiskott-lab/image-reconstruction-mamba}}.

\section{Results}
\subsection{Quantifying Performance}
\label{sec:quantifying_performance}
We first evaluated the performance of trained Mamba variants on augmented Omniglot character images. Specifically, we tested models with different architectural settings, including the number of Mamba blocks, state dimensionality, and training sequence lengths $T_I$ and $T_Q$ (the number of image and query tokens during training). At evaluation time, we varied the number of image and query tokens provided to the model, denoted $V_I$ and $V_Q$. Performance was measured as the MSE between the $V_Q$ reconstruction tokens, queried at random spatial locations, and their corresponding ground-truth patches, given $V_I$ image tokens sampled from random locations.

Additionally, we compared our model against a transformer baseline configured under comparable conditions, i.e., identical token dimensions and varying $T_I$, $T_Q$, and model depth. The transformer encoder followed a Vision Transformer-style architecture~\cite{dosovitskiy_image_2020}, omitting the class token. A transformer decoder processed query tokens via cross-attention to the encoder's image tokens (see the accompanying repository for implementation details).

\begin{figure}[h] 
    \centering 
    \includegraphics[width=1.0\linewidth]{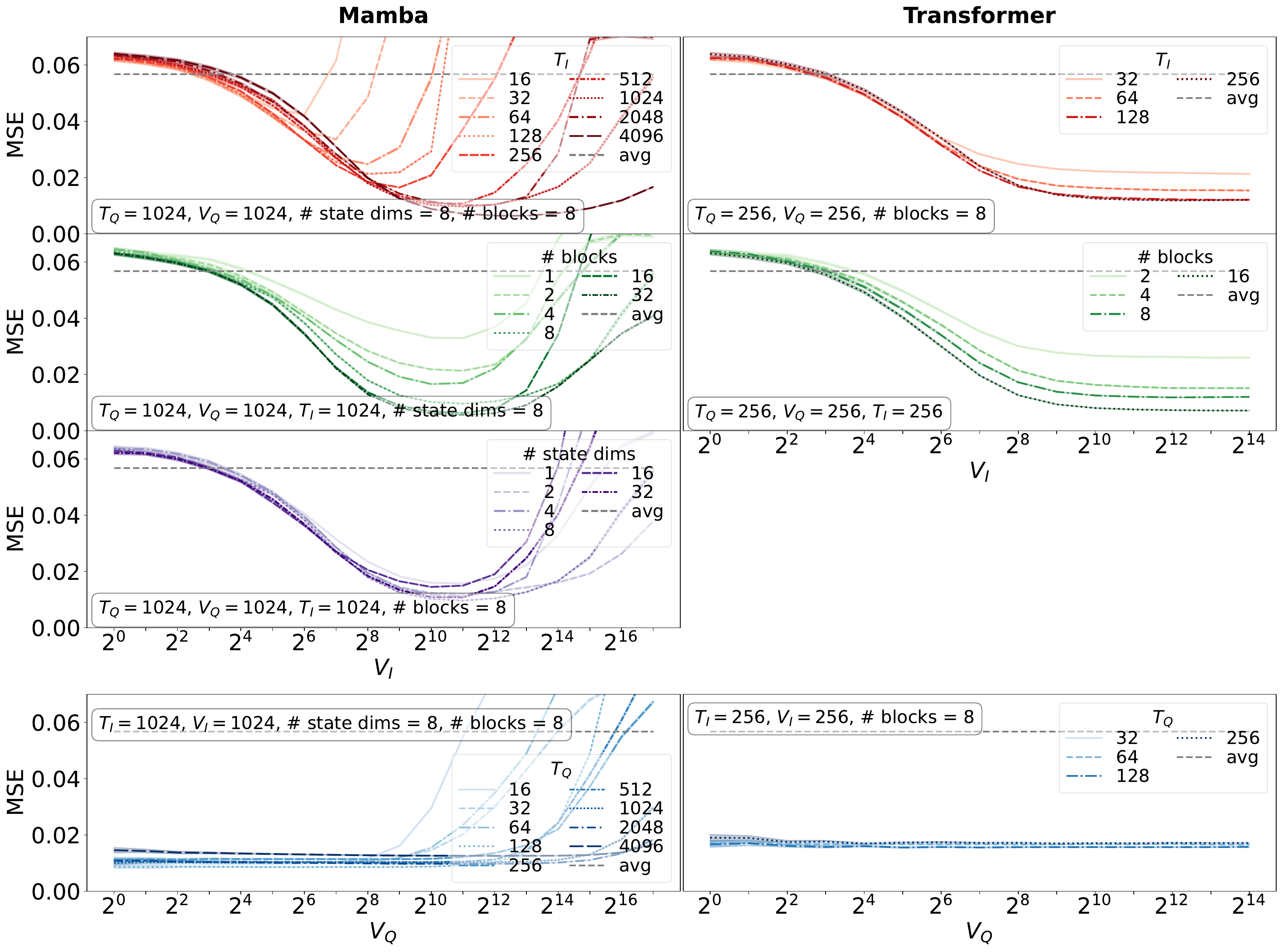}
    \caption{Performance of Mamba \textbf{(Left)} and Transformers \textbf{(Right)} as a function of image and query token sequence lengths under different architectural settings. Curves show mean MSE on the Omniglot test set with 95\% confidence intervals. Avg denotes a baseline that predicts the dataset-wide mean image patch.} 
    \label{fig:curves} 
\end{figure}

\Cref{fig:curves} summarizes the results. For Mamba (Left), the training image-token length $T_I$ strongly influenced length generalization. Naturally, performance increased with increasing $V_I$ as the models gained more knowledge of the input image. The best performances occurred when $V_I \approx 2T_I$. However, for larger values (e.g., $V_I \approx 4T_I$), reconstruction quality deteriorated sharply and quickly fell below chance. Interestingly, models trained with larger $T_I$ values performed slightly worse on shorter sequences than those trained with smaller $T_I$.

Increasing model depth substantially improved overall performance but did not affect length generalization: performance consistently collapsed once $V_I$ exceeded roughly $4T_I$. State dimensionality had a smaller impact. Small state sizes slightly degraded performance for $V_I \leq T_I$, but larger states did not consistently improve results or alter the generalization behavior. When varying $T_Q$ (evaluated over $V_Q$ in the bottom-left plot), performance again degraded once $V_Q$ exceeded the training length. However, this effect occurred much later, around $V_Q \approx 16T_Q$, compared to the $4T_I$ threshold observed for image tokens. As with $T_I$, models trained with larger $T_Q$ values performed slightly worse on short sequences, although the effect was less pronounced.

Transformer models exhibited different behavior. Due to the quadratic complexity of transformers, both training and evaluation were performed with shorter sequence lengths. As with Mamba, increasing $T_I$ improved reconstruction quality. However, unlike Mamba, performance did not deteriorate when $V_I$ substantially exceeded $T_I$. Instead, improvements saturated as $V_I$ increased. Furthermore, the tendency of models trained with small $T_I$ to perform slightly better on short sequences was much less pronounced than for Mamba. Increasing transformer depth improved reconstruction performance but did not affect length generalization. Similarly, varying $T_Q$ showed no meaningful dependence on sequence length: Models achieved comparable performance across all $V_Q$ values.

Several observations are noteworthy. Firstly, consistent with conflicting results in the literature, Mamba’s length generalization appears to be task-dependent: The model generalizes much better to increases in $V_Q$ (querying information) than to increases in $V_I$ (incorporating new image patches). We hypothesize that this difference occurs because query tokens do not introduce new information into the hidden state, making them easier to process.

% Specifically, the model can stop updating the hidden state after seeing the separator token and simply start compiling and outputting the queried information for each token individually. There is no need to keep past query tokens in memory. In contrast, each input token can potentially be equally valuable due to limited memory and the uncertainty of what comes next, making it  more difficult to decide if input tokens may be ignored. An ``ideal'' solution that comes to mind could be that the model develops a strategy for detecting when the whole image has been seen so that it can ignore all subsequent input tokens from there on. However, our results suggest that the model instead resorts to the local optimum of directly adapting to the expected amount of input tokens. Under these circumstances, the model integrates new information into its state as fast as necessary to provide good answers when being queried, but not more than necessary for each token to make the best use of the limited memory capacity. 

Secondly, transformers exhibit strong length generalization despite frequently being reported to struggle with increasing sequence length in NLP tasks. We attribute this difference to our experimental setup, in which the model encounters all positional encodings during training. The strong generalization observed, therefore, supports prior work identifying positional encodings as a key bottleneck for transformer length generalization~\cite{kazemnejad_impact_2023}

Thirdly, the number of state dimensions had no observable effect on length generalization. Hence, we were unable to reproduce the results of Chen et al.~\cite{chen_stuffed_2024}, who argued that an overparameterized state space causes failures in length generalization. 

Finally, the performance curves of Mamba as a function of $T_I$ closely resemble those reported by Ruiz and Gu~\cite{ruiz_understanding_2025} for Mamba-2 trained with different context lengths and evaluated on NLP tasks. This similarity suggests that our findings may extend beyond the image domain. Consistent with their observations, we identify the training context length ($T_I$ or $T_Q$) as the primary factor influencing length generalization. For the remainder of our study, we focused on $T_I$, as our model is most sensitive to this parameter, and we expected reconstructions to remain reliable for query tokens within a reasonable range.
  
\subsection{Probing Visualizations} 
\begin{figure}[h] 
    \centering 
    \includegraphics[width=1.0\linewidth]{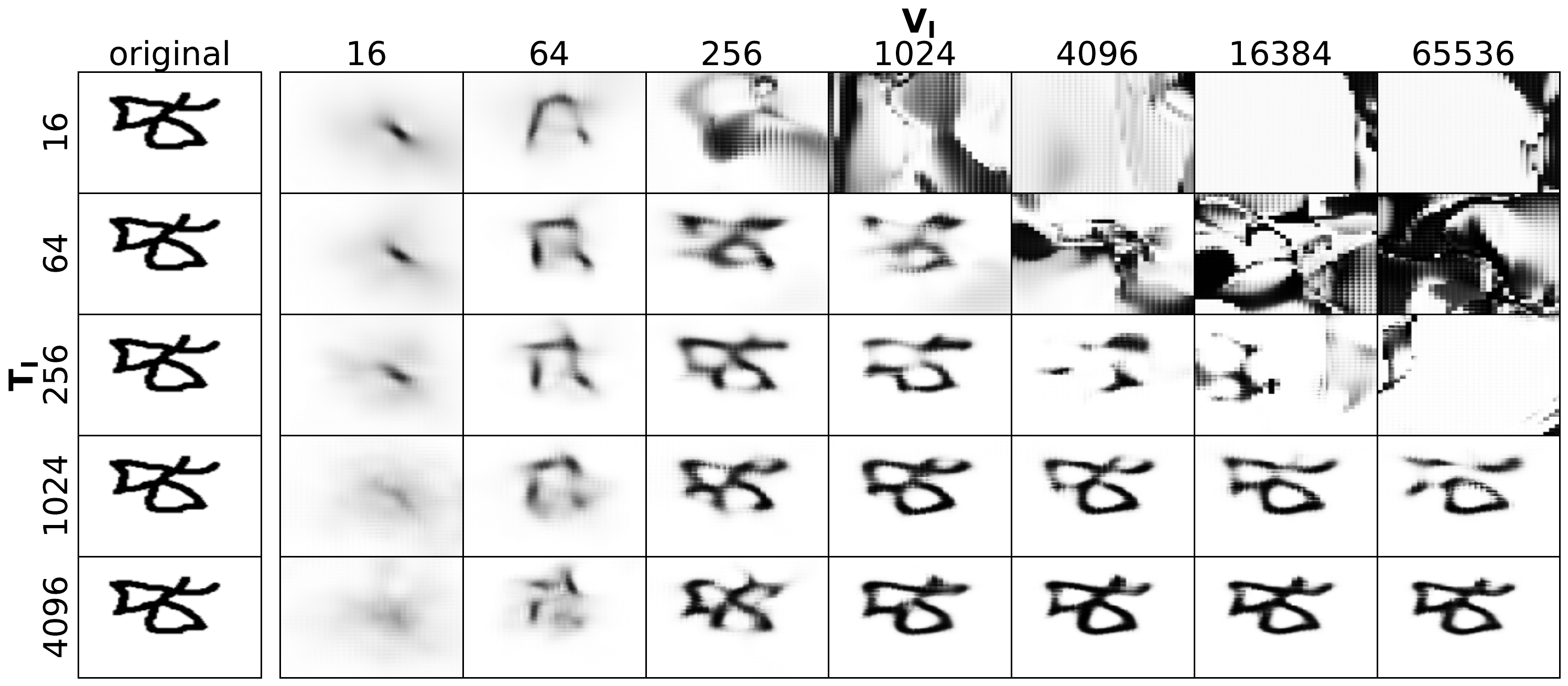}
    \caption{Reconstructions across models. Column 1 shows the original image. Columns 2–8 show reconstructions given the indicated number of randomly sampled image tokens. Rows correspond to models trained with different $T_I$ values.} 
    \label{fig:img_overview} 
\end{figure}

Following the broad performance evaluation, we investigated the influence of $T_I$ using the visualizations enabled by our setup. First, we provided models trained with different $T_I$ values with varying numbers of identical image tokens and visualized the resulting reconstructions (\Cref{fig:img_overview}).

Consistent with our performance evaluation, reconstruction quality improved as $V_I$ increased. Beyond a certain point, however, each model exhibited a sharp degradation after which the original image was no longer recognizable. This breakdown occurred earlier for models trained with shorter context lengths. Interestingly, although identical image tokens were provided across visualizations, reconstructions sometimes differed systematically between models. For example, reconstructions in Column 2 show a higher-contrast center for models with smaller $T_I$, suggesting systematic differences in how the models process early observations.

\begin{figure}[h] 
    \centering 
    \includegraphics[width=1.0\linewidth]{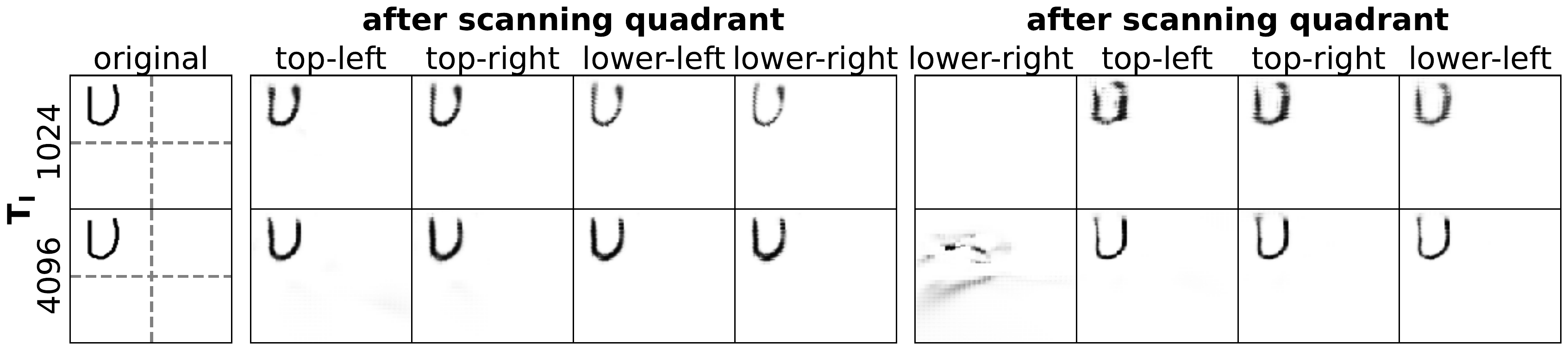}
    \caption{Scanning an image in different orders. The left panel shows the original image. The middle and right panels display reconstructions of the image quadrants with different processing orders. Rows correspond to models with different $T_I$ values.} 
    \label{fig:comb_4_squares} 
\end{figure}

To further investigate insinuated differences, we conducted two probing experiments. In the first experiment, we evaluated models on manipulated images in which the character appeared only in the top-left quadrant. We then processed random sets of image tokens from each quadrant using two scanning orders. In the first order, the top-left quadrant was processed first; in the second order, it was processed second (see \Cref{fig:comb_4_squares}).

When the quadrant containing the character was processed first, both models reconstructed it accurately. However, for the model with $T_I=1024$, the character gradually faded over time, whereas this effect was minimal for the model with $T_I=4096$. Changing the scanning order significantly altered reconstructions: The model with $T_I=1024$ produced blurrier characters, whereas the model with $T_I=4096$ produced weaker contrast.

\begin{figure}[h] 
    \centering 
    \includegraphics[width=1.0\linewidth]{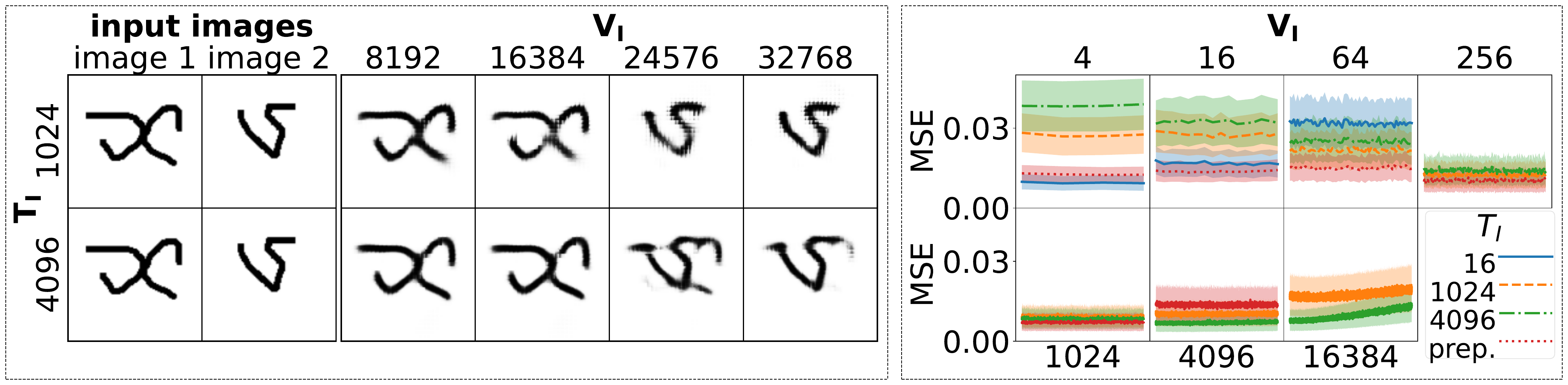}
    \caption{(\textbf{Left).} Sequential processing of two images. Columns 1–2 show the original images. Columns 3–4 show reconstructions after processing tokens from the first image. Columns 5–6 show reconstructions after processing tokens from the first image, followed by those of the second image. Rows correspond to models trained with different $T_I$. \textbf{(Right).} Mean MSE of reconstruction tokens queried at spatial locations corresponding to the provided $V_I$ tokens on the Omniglot test set (95\% confidence intervals). Lines correspond to models trained with different $T_I$ values and a sequence-length–adaptive variant (prepend Mamba).} 
    \label{fig:comb_two_imgs_delay} 
\end{figure}

In the second probing experiment, we examined how models respond to changes in the input image. Specifically, the model processed 16,384 tokens from one image, followed by 16,384 tokens from a second image, giving the model sufficient time to adapt to the second image. We generated reconstructions at intermediate steps (\Cref{fig:comb_two_imgs_delay} (Left)). Both models eventually adapted to the second image after observing sufficient tokens. However, the model trained with $T_I=1024$ adapted much more rapidly: After processing 8192 tokens from the second image, its reconstruction reflected only the second image. In contrast, the model with $T_I=4096$ still retained artifacts from the first image at the same point.

Both probing experiments indicate qualitative differences in the models' processing dynamics, particularly in how quickly they incorporate and update information over time. To support these qualitative observations with quantitative evidence, we measured how the MSE evolves during sequence processing for individual patches. Specifically, we queried only locations corresponding to image tokens whose spatial positions had already been observed and tracked how the reconstruction error at these locations changed over time. In this setting, the model already has full information about the queried patches, meaning that the MSE reflects only the model's dynamics rather than additional information.

Both probing experiments illustrate qualitative differences in the models' processing dynamics, particularly in how quickly they incorporate and update information over time. To quantify these observations, we measured how the MSE evolves during sequence processing for individual patches. Specifically, we queried only spatial locations corresponding to image tokens whose positions had already been observed. We tracked how the reconstruction error at these locations changed as additional tokens were processed. In this setting, the model already has full information about the queried patches, so the MSE reflects only the model's internal dynamics rather than the availability of new information.

Results are shown in \Cref{fig:comb_two_imgs_delay} (Right) for models trained with three distinct $T_I$ values and a modified Mamba variant called prepended Mamba, which we introduce in detail in the next section. The model trained with a small $T_I$ value achieves low MSE for patches observed early in the sequence but fails to maintain this performance as additional tokens are processed. In contrast, models trained with larger $T_I$ values initially exhibit higher MSE, but their performance improves as more tokens are processed, even for patches that have already been observed. This pattern suggests that although information from early image tokens is incorporated into the hidden state early in the sequence, it becomes accessible to query tokens only at later stages, unlike in models with small $T_I$, indicating a strong influence of $T_I$ on the model’s processing speed.

The illustrated qualitative differences in processing observed for models trained with different $T_I$, supported by quantitative evidence, suggest that Mamba adapts its computational strategy to the sequence lengths encountered during training. Moreover, together with the performance evaluation in \Cref{sec:quantifying_performance}, which shows that models trained with small $T_I$ perform better on short sequences than those with large $T_I$, our results indicate that Mamba implicitly learns to trade off performance across the sequence length regimes encountered during training.

% The visualizations provided further insight into the behavior of the models. Although all information about the top-left quadrant was already available after it had been scanned, reconstruction quality continued to improve even as the model processed unrelated patches from the remaining quadrants. Taking \Cref{fig:img_overview} into account, this observation suggests that the longer-context model processes image information slowly over multiple time steps compared to models with short $T_I$.

\subsection{Adapting Mamba}
\label{sec:adapting_mamba}
The observation that Mamba learns to trade off performance across sequence length regimes leads to several predictions. Firstly, restricting training to the upper half of the interval should improve performance within that range. To test this hypothesis, we trained a model with image token sequences lengths restricted to 512–1024 (referred to as truncated Mamba). Secondly, increasing the training interval should require sacrificing performance on short sequences to achieve better performance on longer ones. We tested this prediction by retraining our default model with $T_I=65536$ (expanded Mamba). Finally, because Mamba does not know the sequence length in advance, it cannot explicitly adapt its dynamics to varying sequence lengths and instead learns behavior that performs well on average across training lengths. To enable length-dependent adaptation of Mamba, we prepend a token encoding the sequence length to the input sequence (prepended Mamba), allowing the model to store this information in its hidden state and adjust its processing accordingly.

\begin{figure}[h] 
    \centering 
    \includegraphics[width=1.0\linewidth]{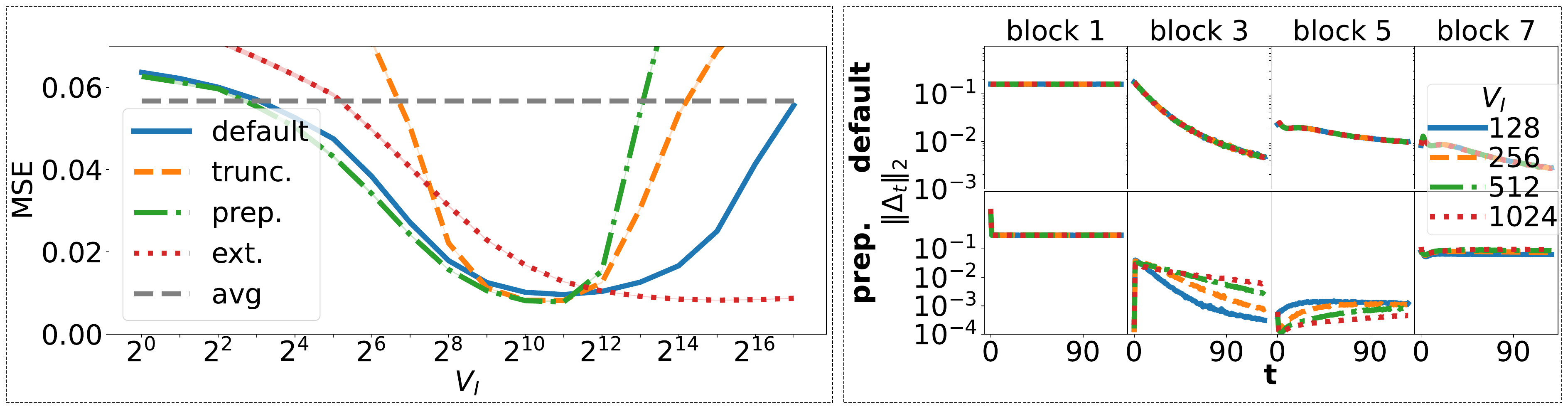}
    \caption{\textbf{(Left).} Mean MSE of adapted Mamba variants across varying image-token sequence lengths, evaluated with 1024 query tokens on the Omniglot test set, with 95\% confidence intervals. \textbf{(Right).} Each column corresponds to a representative block of the model. The top row shows the default model, and the bottom row shows the prepended Mamba variant. The curves depict the evolution of $\|\mathrm{\Delta}_t\|_2$ over the first 128 tokens, indexed by $t$, for different $V_I$ values, averaged across test sequences.}
    \label{fig:comb_conc_plot} 
\end{figure}

We report the performance of the adapted trained models in \Cref{fig:comb_conc_plot} (Left), following the same evaluation protocol as in \Cref{sec:quantifying_performance}. As expected, truncated Mamba achieves slightly better performance within its training interval, though it degrades more sharply than the default model outside this range. As hypothesized, expanded Mamba sacrifices performance on short sequences while improving substantially on longer ones. The length-adaptive variant outperforms the default model across sequence lengths within $T_I$. This difference becomes particularly evident when querying only previously observed locations (see \Cref{fig:comb_two_imgs_delay} (right)), where prepended Mamba performs significantly better on short sequences than the default model with $T_I = 1024$. However, prepended Mamba also reduces length generalization compared to the default model.

To verify that prepended Mamba indeed demonstrates length-adaptive behavior, we analyzed the evolution of the Euclidean norm $\|\mathrm{\Delta}_t\|_2$ of the default model and prepended Mamba across blocks, averaged over test sequences of different lengths. Rather than interpreting absolute values, we examine whether the trajectory of $\|\mathrm{\Delta}_t\|_2$ over image tokens varies with total sequence length, which would indicate adaptive hidden-state dynamics. Results for the first 128 tokens are shown in \Cref{fig:comb_conc_plot} (right). For prepended Mamba, $\|\mathrm{\Delta}_t\|_2$ evolves differently across sequence lengths, whereas in the default model it remains nearly invariant. Note that the average $\|\mathrm{\Delta}_t\|_2$ for the first block is constant by design when tokens are drawn from the same distribution, since at this stage it depends only on the incoming token values and not on the hidden state. For prependend Mamba, $\|\mathrm{\Delta}_0\|_2$ differs because the first token is drawn from the distribution of sequence-length tokens rather than from the distribution of image tokens.

Our evaluation of the adapted Mamba variants provides additional evidence that Mamba trades off performance across the sequence lengths encountered during training: Training on longer sequences leads to significantly worse performance on shorter sequences, whereas restricting training to narrower length ranges or enabling sequence-length–adaptive behavior improves performance within the trained interval. However, the specializations also reduce length generalization.

\section{Discussion}
In this work, we set out to study length generalization in Mamba using a purposefully designed image reconstruction task that enables targeted probing of the model’s processing. 

Our results suggest that Mamba does not learn a sequence-length–agnostic algorithm. Instead, it adapts its computational strategy to the sequence lengths encountered during training, particularly by adjusting processing speed. These adaptations lead to distinct behaviors across sequence-length regimes that do not transfer beyond the training distribution. In effect, the model optimizes a task-dependent trade-off across the sequence lengths observed during training. In contrast, transformers exhibit strong length generalization in our setting, likely because they are not required to extrapolate to unseen positional encodings.

The trade-off observed in our experiments raises the question of how related approaches (\Cref{sec:related_work}) achieve improved length generalization. The methods proposed in DeciMamba~\cite{ben-kish_decimamba_2024} and LongMamba~\cite{ye_longmamba_2025} both rely on omitting tokens deemed unimportant during processing. Although these works demonstrate improved length generalization, Lu et al.~\cite{lu_mamba_2025} report that both methods perform worse than the default Mamba model on sequences of around 2000 tokens and outperform it only for longer sequences ($\geq$4000 tokens). We expect this performance gap to be even larger for shorter sequences. In this sense, these methods are consistent with our findings, suggesting a trade-off across sequence-length regimes: Improvements on long sequences may come at the cost of degraded performance on shorter ones.

In contrast to DeciMamba and LongMamba, MambaExtend~\cite{azizi_mambaextend_2025} and MambaModulation~\cite{lu_mamba_2025} report improved performance over the base model on both short and long sequences (although the evaluated sequences are not shorter than 2000 tokens). Importantly, their approaches differ from the former methods in that they introduce sequence-length–dependent scaling parameters. Therefore, their results are consistent with ours, which show that length-adaptive Mamba variants can alleviate the observed trade-off. 

Sequence-length–adaptive mechanisms appear promising, but they also introduce an important limitation. Whether implemented by explicitly providing the sequence length at the beginning of the sequence, as in our approach, or through scaling mechanisms such as MambaExtend or MambaModulation, these methods require the sequence length to be known in advance. In many applications, however, the sequence length may not be known before processing begins. Nevertheless, our results also suggest that Mamba performs well across a range of sequence lengths, indicating that a rough estimate of the sequence length may already be sufficient.

% From a broader perspective, our method provides an intuitive framework for studying length generalization in Mamba, it is currently limited to the vision domain. However, the similarity between our performance curves across sequence lengths and those reported by Ruiz and Gu~\cite{ruiz_understanding_2025} for a comparable analysis in language modeling suggests that these effects may generalize beyond vision. Future work should therefore investigate whether similar phenomena arise in NLP, the primary application domain of Mamba, in the literature.

Taken together, our results suggest that length generalization in Mamba is constrained by a trade-off in performance across sequence lengths. The model adapts to the sequence-length distribution encountered during training rather than learning a length-invariant algorithm. Understanding and controlling this trade-off may therefore be key to developing Mamba variants that maintain strong performance across a broader range of sequence lengths. In this context, sequence-length–adaptive mechanisms appear to be a particularly promising direction.

\begin{credits}
\subsubsection{\ackname} 
This work was partially supported by the RUB Research School through its PR.INT program.

\subsubsection{\discintname}
The authors have no competing interests to declare that are
relevant to the content of this article.
\end{credits}
\bibliographystyle{splncs04}
\bibliography{references}

\end{document}